\documentclass[journal]{IEEEtran}

\usepackage{times}
\usepackage{helvet}
\usepackage{courier}
\usepackage[hyphens]{url}
\usepackage{graphicx}
\usepackage[nocompress]{cite} 
\usepackage{caption}
\usepackage{booktabs}
\usepackage{multirow}
\usepackage{tabularx}
\usepackage{array}
\usepackage{xcolor}
\usepackage[table]{xcolor}
\newcommand{\rev}[1]{\textcolor{black}{#1}}
\usepackage{diagbox}
\definecolor{topone}{RGB}{220,53,69}    
\definecolor{toptwo}{RGB}{0,123,255}    
\definecolor{topthree}{RGB}{40,167,69}  

\newcommand{\TopOne}[1]{\textcolor{topone}{\textbf{#1}}}
\newcommand{\TopTwo}[1]{\textcolor{toptwo}{#1}}
\newcommand{\TopThree}[1]{\textcolor{topthree}{#1}}

\usepackage{dblfloatfix}
\usepackage{placeins}
\usepackage{amsmath,amssymb}
\usepackage{algorithm}
\usepackage{algorithmic}

\usepackage[colorlinks, linkcolor=blue, urlcolor=blue, citecolor=blue]{hyperref}

\begin{document}

\title{BladeYOLO: Wind Turbine Blade Defect Detection with Limited Annotations and Weak-Saliency Awareness}

\author{
Yabin Xu, \textit{Member, IEEE}, Fangtao Zhang, Fan Wang, Zhan Wang, Honghua Chen, \textit{Member, IEEE}, Mingqiang Wei, \textit{Senior Member, IEEE}, Haoran Xie, \textit{Senior Member, IEEE}, and Sam Kwong, \textit{Fellow, IEEE}
\thanks{Y. Xu, F. Zhang, and F. Wang are with the School of Mechanical Engineering, Zhejiang Sci-Tech University, Hangzhou 310018, China (e-mail: yabinxu@zstu.edu.cn; fqiu42610@gmail.com; fwang@zstu.edu.cn).}
\thanks{Z. Wang is with the Department of Artificial Intelligence and Robotics, Zhejiang Energy Digital Technology Co., Ltd, Hangzhou 311100, China (e-mail: wangzhan.robotic@outlook.com).}
\thanks{M. Wei is with the School of Computer Science and Technology, Nanjing University of Aeronautics and Astronautics, Nanjing, China (e-mail: mingqiang.wei@gmail.com).}
\thanks{H. Chen, H. Xie, and S. Kwong are with Lingnan University, Hong Kong SAR, China (e-mail: honghuachen@ln.edu.hk; hrxie@ln.edu.hk; samkwong@ln.edu.hk).}
\thanks{Corresponding author: Honghua Chen.}
}

\IEEEtitleabstractindextext{%
\begin{abstract}
Wind turbine blade defect detection remains highly challenging in real-world inspection scenarios due to limited on-site data and the subtle visual characteristics of defects. In practice, blade defects are often small-scale, low-contrast, and difficult to distinguish from complex backgrounds, which significantly limits the robustness of existing detectors.
To address these challenges, we propose \emph{BladeYOLO}, a defect detection framework for wind turbine blades. Specifically, we integrate a Vision Transformer (ViT) backbone initialized with DINOv3 self-supervised pre-trained weights into YOLOv12-L, enabling the transfer of large-scale generic visual priors to blade defect detection and improving feature representation under limited training annotations. To enhance the perception of subtle defects, we further develop a \emph{Mamba-guided Weak-Defect Enhancement} module, which consists of a \emph{Detail-Enhanced Multi-scale Branch} for preserving high-frequency structural cues and a \emph{Cross-Mamba} module for progressively propagating high-level semantic guidance to shallow features. In addition, we introduce a lightweight \emph{Style-Injector} module that captures environment-related style information via Fourier decomposition and injects it into selected ViT self-attention layers, thereby improving robustness against environment-induced appearance variations. \textcolor{black}{Extensive experiments demonstrate that BladeYOLO achieves superior performance on the WTBlade-Defect dataset, with additional annotation-budget experiments showing its favorable performance under reduced training annotations. Evaluation on the public Wind Surface Defect dataset further provides supportive evidence for the cross-dataset robustness of BladeYOLO. In particular, on this public dataset, BladeYOLO outperforms the best competing method by 3.5\% in mAP$_{50}$ and 2.5\% in mAP$_{50-95}$. The code is available at \url{https://github.com/zhangfangtao/BladeYOLO}.}


%

\end{abstract}

}

\maketitle
\IEEEdisplaynontitleabstractindextext
\IEEEpeerreviewmaketitle


\section{Introduction}
\label{sec:intro}
Accurate detection of surface defects on wind turbine blades is essential for the safe and reliable operation of wind power systems~\cite{herbert2007review}. With the rapid development of deep learning and computer vision, vision-based inspection has become an attractive solution due to its high efficiency, low cost, and reduced dependence on manual labor~\cite{roga2022recent}. Existing vision-based methods are mainly built upon either two-stage or one-stage detectors~\cite{Du2020Overview}. Two-stage detectors, such as Faster R-CNN~\cite{Chen2021Scale} and Mask R-CNN~\cite{he2017mask}, usually achieve strong accuracy by first generating region proposals and then performing classification and localization on region-of-interest features. However, their multi-stage pipeline incurs considerable computational overhead, limiting their suitability for efficient real-world blade inspection~\cite{chen2016r}. In contrast, one-stage detectors represented by the YOLO family~\cite{Jiang2022Review,chen2024gcn,fang2019tinier} directly predict object categories and bounding boxes in a single forward pass, offering a more practical trade-off between accuracy and efficiency.

Despite this progress, wind turbine blade defect detection remains far from solved in real inspection scenarios. The core difficulty lies in two aspects. First, on-site training data are often scarce, making it difficult for existing detectors to learn robust and transferable representations. Second, blade defects often exhibit weak visual saliency: they are typically small-scale, low-contrast, and easily overwhelmed by background textures. These challenges are further aggravated by environmental variations, such as illumination changes and foggy or hazy conditions, which degrade image quality and suppress already subtle defect cues. As a result, existing detectors often suffer from unstable predictions and missed detections under practical inspection conditions.

Recent studies have attempted to improve blade defect detection by enhancing YOLO-based architectures. For example, Zhu et al.~\cite{Zhu2022Research} incorporated MobileNetV3~\cite{howard2019searching} and GhostNet~\cite{han2020ghostnet} into YOLOv5~\cite{jocher2022ultralytics} to strengthen feature extraction. 
\rev{Zhang et al.~\cite{zhang2024attention} introduced a YOLOv8-based wind turbine blade defect detection method that incorporates attention mechanisms and multi-scale feature fusion to improve defect representation under complex backgrounds.} 
Xu et al.~\cite{Xu2025Wind} proposed SNMSDA-YOLO11 with multi-scale dilated attention for multi-type defect detection. 
Although these methods improve performance to some extent, they still operate largely within the conventional supervised detection paradigm and mainly rely on architectural refinement, attention-based enhancement, and multi-scale feature modeling. Consequently, their effectiveness remains strongly dependent on sufficient training data and adequate environmental coverage during training. Under scarce data, degraded imaging conditions, and weak defect cues, these methods still struggle to achieve stable and reliable detection.

\begin{figure}[thbp]
    \centering \includegraphics[width=1.0\columnwidth]{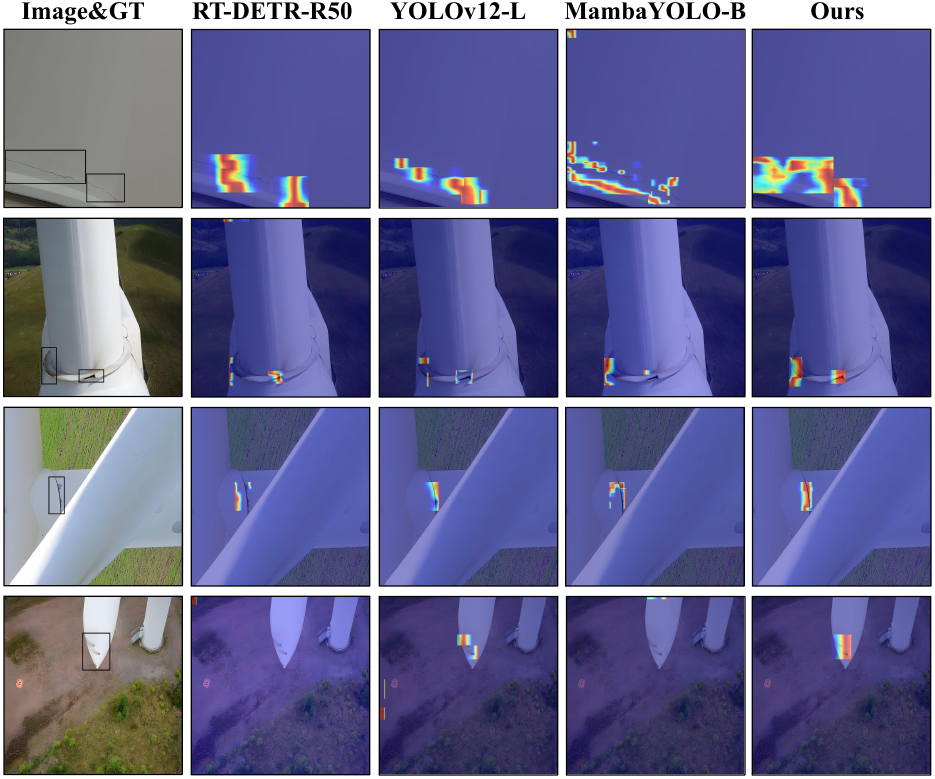}
   \caption{Heatmap-based qualitative comparison of different detectors on wind turbine blade defect detection. From left to right are the input image with ground truth, RT-DETR-R50, YOLOv12-L, MambaYOLO-B, and the proposed BladeYOLO. Compared with the competing methods, BladeYOLO produces more accurate and concentrated responses on defect regions while suppressing background interference more effectively, especially in challenging cases involving complex backgrounds and weak-saliency defects.}
    \label{fig:fig_1}
\end{figure}

These limitations suggest that improving blade defect detection under limited-data and weak-saliency conditions requires more than incremental detector refinement. Our key insight is that the problem should be addressed from three complementary aspects. First, stronger transferable representation priors are needed to reduce the dependence on large-scale task-specific training data. Second, subtle defect cues should be explicitly preserved and enhanced, rather than being passively handled by generic multi-scale features. Third, shallow features should receive stronger semantic support to improve the perception of small and low-saliency defects. In addition, because environmental variations further distort global appearance and suppress fine-grained details, the transferred representations should be adaptively adjusted to remain robust across diverse inspection conditions.

Fortunately, large-scale self-supervised visual models provide a promising foundation for this purpose~\cite{liu2025robust}. In particular, DINOv3~\cite{simeoni2025dinov3}, pre-trained on massive web-scale image data, learns rich and transferable visual representations and has demonstrated strong generalization across a wide range of downstream tasks~\cite{Yang2025Segdino, Huang2025Real-Time, yuan2025ad}. This progress opens new opportunities for industrial detection. Inspired by these advances, we seek to transfer the representation priors of DINOv3 to wind turbine blade defect detection, so as to alleviate the performance bottleneck caused by scarce training data and provide a stronger feature foundation for robust detection.

Based on the above insights, we propose \emph{BladeYOLO}, a wind turbine blade defect detector designed for limited-data settings and weak-saliency awareness. As illustrated by the heatmap-based qualitative comparison in \autoref{fig:fig_1}, BladeYOLO produces more concentrated activations on defect regions, whereas existing detectors often generate scattered responses on low-contrast targets. This qualitative evidence suggests that the integration of DINOv3 representation priors and Mamba-guided calibration helps distinguish subtle defect cues from background interference. Specifically, we adopt YOLOv12-L as the detection framework and replace its original backbone with a Vision Transformer (ViT) initialized with DINOv3 pre-trained weights, enabling the transfer of large-scale self-supervised visual knowledge to blade defect detection. To improve the perception of small and weakly salient defects, we further develop a \emph{Mamba-guided Weak-Defect Enhancement (MWE)} module, which consists of two components: a \emph{Detail-Enhanced Multi-scale Branch} for explicitly preserving high-frequency structural cues, and a \emph{Cross-Mamba} module for progressively propagating high-level semantic guidance to shallow features. In this way, subtle defect details are first enhanced and then semantically calibrated in a coarse-to-fine manner, thereby reducing missed detections of small and low-saliency defects. To further improve robustness against environment-induced appearance variations, we utilize a \emph{Style-Injector} (SI) module that extracts style-related information through Fourier decomposition and injects it into selected self-attention layers of the DINOv3-initialized ViT using lightweight learnable branches~\cite{qian2011algorithm,chen2022adaptformer}.
\rev{Extensive experiments on the WTBlade-Defect dataset and the public Wind Surface Defect dataset demonstrate that BladeYOLO consistently outperforms existing methods and shows promising generalization performance. Dedicated evaluations under weak-saliency and limited-annotation settings further highlight the distinctive advantages of BladeYOLO in weak-defect perception and annotation-efficient defect detection.
}

The main contributions are summarized as follows:
\begin{enumerate}
    \item \rev{We propose \emph{BladeYOLO}, a YOLOv12-L-based defect detection framework for wind turbine blades that integrates a DINOv3-initialized ViT backbone, enabling the transfer of large-scale self-supervised visual priors to improve defect representation under data-scarce conditions.}
    
    \item \rev{We develop an \emph{MWE} module for improving the perception of small and low-saliency defects. It consists of a Detail-Enhanced Multi-scale Branch, which explicitly preserves high-frequency structural cues, and a Cross-Mamba module, which progressively propagates high-level semantic guidance to shallow features for semantic calibration of weak defects}.
    
    \item \rev{We introduce a lightweight \emph{Style-Injector} module that explicitly captures environment-related style information and injects it into selected ViT self-attention layers, thereby enabling environment-aware feature adaptation to appearance variations such as illumination changes and foggy or hazy conditions.}
\end{enumerate}

\section{Related Work}
\label{sec:related}
\subsection{Generic Object Detection Models}

\rev{Generic object detection provides a fundamental basis for wind turbine blade defect detection. Deep learning-based detectors have significantly improved accuracy and efficiency and are categorized into two-stage and one-stage methods.}

\rev{Two-stage detectors first generate region proposals and then perform classification and localization on region-of-interest features. Representative methods include Fast R-CNN~\cite{girshick2015fast}, Faster R-CNN~\cite{ren2015faster}, Cascade R-CNN~\cite{cai2018cascade}, and TridentNet~\cite{li2019scale}. These methods achieve high accuracy through proposal-based refinement, but their multi-stage design incurs considerable computational overhead, limiting their suitability for efficient real-world blade inspection.}

\rev{In contrast, one-stage detectors directly predict object categories and bounding boxes in a single forward pass, providing a better trade-off between accuracy and efficiency. SSD~\cite{liu2016ssd} introduced multi-scale prediction strategies, while YOLO~\cite{redmon2016you} formulated object detection as a single regression problem and became a representative real-time detector. Later YOLO-based detectors, such as YOLOv7~\cite{wang2023yolov7} and YOLOv10~\cite{wang2024yolov10}, further improved detection performance through effective network structures, feature fusion designs, and detection heads. In addition, anchor-free detectors, such as CornerNet~\cite{law2018cornernet}, avoid predefined anchors and provide a simpler detection design.}

\rev{Due to their favorable trade-off between accuracy and efficiency, one-stage detectors serve as suitable baselines for wind turbine blade defect detection. However, generic detectors may not fully adapt to real-world blade inspection scenarios. Task-specific feature enhancement and effective semantic guidance are still needed for robust blade defect detection.}

\subsection{Wind Turbine Blade Defect Detection}

\rev{Wind turbine blade defect detection has attracted increasing attention in inspection tasks~\cite{song2024topology,sun2025improved}. In complex engineering scenarios, reliable analysis remains essential. In practice, blade defects are often small, low-contrast, and easily obscured by complex backgrounds, while environmental variations require detectors to be accurate, robust, and efficient.}

\rev{Existing studies have improved blade defect detection from three perspectives, including multi-scale feature modeling, attention-based feature enhancement, and lightweight network design. He et al.~\cite{he2024adaptive} proposed an adaptive detection method to improve the detection of defects at different scales through multi-level feature fusion and a multi-scale proposal module. Yang et al.~\cite{yang2025research} developed FRE-DETR by enhancing the FasterNet~\cite{chen2023run} backbone and introducing a lightweight feature selection and fusion module for complex defect detection. Ji et al.~\cite{ji2025cmfcanet} proposed CMFCANet, which enhances the recognition of blade defects in low-light environments through the RGB-HSI module and channel attention mechanism. Zhao et al.~\cite{zhanfang2025enhancing} improved YOLOv8 by adding an extra small-object detection layer to enhance the perception of tiny blade defects. Liu et al.~\cite{Liu2025windturbine} designed a lightweight CAM-DW module in YOLOv10 to strengthen feature fusion while maintaining efficiency. Ma et al.~\cite{ma2024wind} proposed MES-YOLOv8n, which improves the detection of tiny defects through the MC2f module and ECA attention mechanism.}

\rev{Overall, existing methods have made clear progress in multi-scale modeling, local feature enhancement, and lightweight deployment. For low-saliency targets under complex backgrounds, previous studies have shown that multi-scale attention, two-stream pyramid modeling, nested feature aggregation, and refinement mechanisms can improve target representation and background suppression~\cite{cong2021rrnet,li2019nested,cong2023point}. However, most of them still improve performance mainly through detector architecture refinement within a conventional supervised learning pipeline. When training data are scarce and defect cues are weak, these methods remain vulnerable to unstable predictions, especially under complex inspection conditions with strong background interference and large defect-scale variations. Therefore, further improvements require not only more effective detector structures but also more transferable representation priors and more effective weak-defect perception mechanisms.}

\subsection{Visual Foundation Models for Defect Detection}

\rev{Compared with conventional supervised methods that rely on labeled data, visual foundation models are pre-trained on data collections and can provide transferable representation priors for downstream tasks. Such priors are valuable for defect detection when task-specific training data are scarce and defect cues are weak, helping reduce the dependence on task-specific annotations and improve feature representation.}

\rev{Self-supervised pre-training is an important way to learn transferable visual representations. Representative methods include MoCo~\cite{he2020momentum}, which improves contrastive learning through a momentum encoder; SimSiam~\cite{chen2021exploring}, which learns effective representations without negative samples; and DINO~\cite{caron2021emerging}, which enables Vision Transformers to learn rich semantic representations via self-distillation without manual annotations. More recently, DINOv3~\cite{simeoni2025dinov3} further improves representation transferability through large-scale self-supervised pre-training.}

\rev{Recent studies have explored pre-trained visual models in industrial defect scenarios. Hu et al.~\cite{hu2023steel} performed contrastive pre-training on unlabeled steel surface images and transferred the learned weights to Faster R-CNN and RetinaNet, improving steel defect detection performance. Xu et al.~\cite{xu2025application} adopted SimSiam for pre-training on unlabeled steel surface images and used the pre-trained model as the feature extractor of Faster R-CNN under limited-data conditions. Nevertheless, most existing pre-trained visual models are developed using natural images or relatively simple industrial surface images, and their learned representations are not optimized for low-saliency defects and environmental variations in complex blade inspection scenarios. Hence, adaptation of pre-trained visual representations to wind turbine blade defect detection remains underexplored, especially under limited-data and weak-saliency conditions~\cite{chen2026empowering}.}

\subsection{Mamba-based Visual Modeling}

\rev{Effective contextual modeling is important for visual recognition and detection tasks. Conventional visual architectures, such as CNNs and Transformers, are used for modeling. CNNs are effective in capturing local patterns but have limited global modeling ability, while Transformers can model long-range dependencies but usually suffer from the high computational cost of self-attention. To balance contextual modeling and computational efficiency, Mamba introduces selective state-space modeling with linear complexity, providing an efficient alternative for long-range feature modeling~\cite{gu2024mamba}.}

\rev{Recent studies have extended Mamba-style modeling to visual tasks. VMamba adapts state-space modeling to two-dimensional visual features through selective scanning~\cite{zhu2024vision}. Following visual state-space modeling, Wu et al.~\cite{wu2025mv} proposed MV-YOLO, which introduces Mamba-based vision modules into a YOLO detection framework for UAV and remote-sensing small-object detection. These studies demonstrate the potential of Mamba-based modeling for enhancing contextual representation in visual perception and detection tasks.}

\rev{However, existing Mamba-based visual methods are oriented toward general visual representation or aerial small-object detection, while their adaptation to weak-saliency industrial defect detection remains limited. This motivates the proposed Mamba-guided Weak-Defect Enhancement (MWE) module, in which Cross-Mamba propagates high-level semantic guidance to shallow features for semantic calibration, enhancing the perception of small and low-saliency defects.}

\begin{figure*}[h!]
\centering\includegraphics[width=1.0\textwidth]{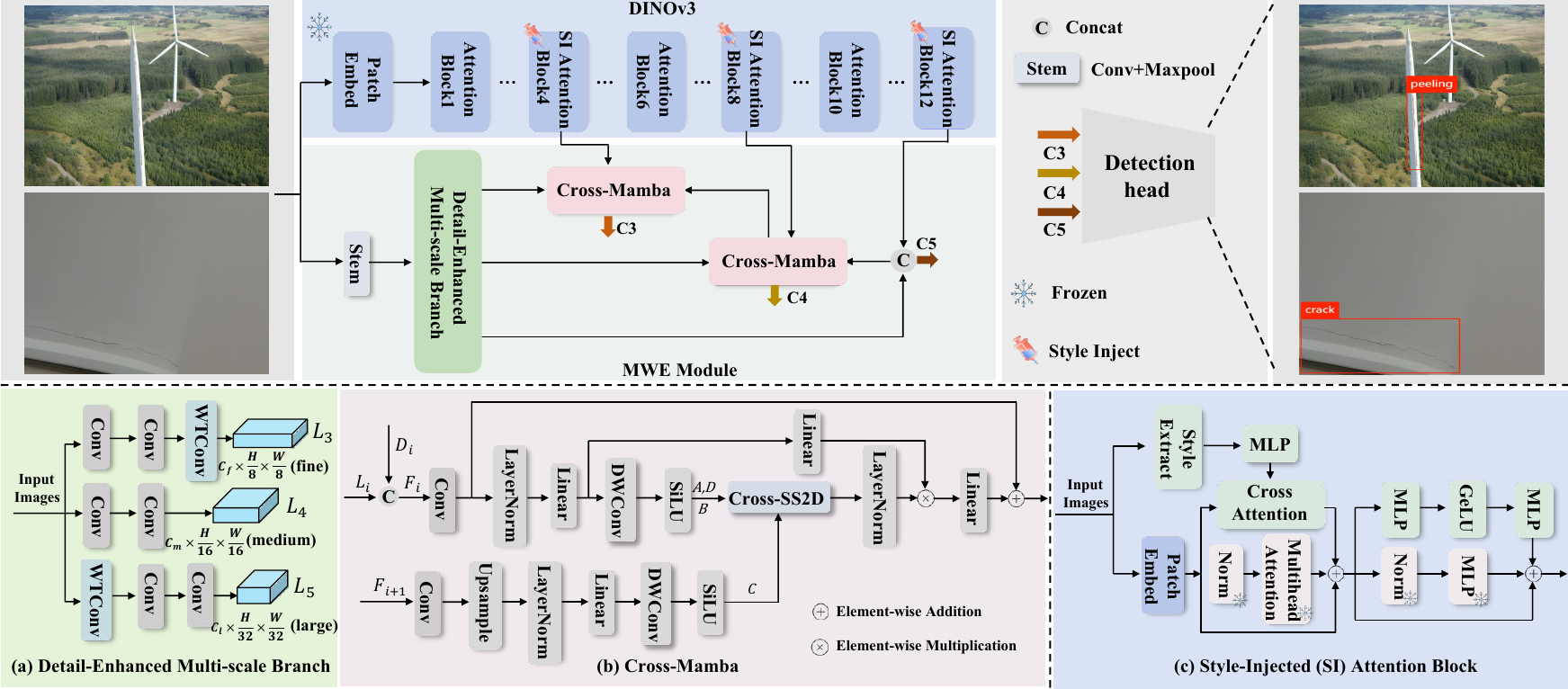} 
    \caption{\rev{Overview of the proposed BladeYOLO framework. The original YOLOv12-L backbone is replaced with a DINOv3-initialized ViT backbone to leverage large-scale self-supervised representation priors. To improve the perception of small and low-saliency defects, we develop a Mamba-guided Weak-Defect Enhancement Module, which consists of a Detail-Enhanced Multi-scale Branch for preserving high-frequency structural cues and a Cross-Mamba module for progressive cross-scale semantic calibration. In addition, a lightweight Style-Injector is sparsely inserted into selected self-attention blocks to improve robustness against environment-induced appearance variations. The resulting multi-level features \{C3, C4, C5\} correspond to the shallow, intermediate, and deep feature maps with spatial resolutions of $\frac{H}{8} \times \frac{W}{8}$, $\frac{H}{16} \times \frac{W}{16}$, and $\frac{H}{32} \times \frac{W}{32}$, respectively, and are then fed into the YOLOv12-L neck and detection head for final defect classification and localization.}}
    \label{fig:fig_2}
\end{figure*}

\section{Method}
\label{sec:method}

\subsection{Motivation and Overall Architecture}

Self-supervised pre-training provides a promising way to alleviate the reliance on large amounts of task-specific training data in industrial defect detection. Benefiting from pre-training on a large dataset, DINOv3~\cite{simeoni2025dinov3} learns strong and transferable representations and has demonstrated competitive performance across a variety of downstream vision tasks. This makes it an appealing foundation for wind turbine blade defect detection, where training data are often scarce. However, directly transferring such pre-trained representations to blade inspection remains challenging. In real inspection scenarios, blade images are frequently affected by illumination variations and foggy or hazy conditions, while defect patterns are often small-scale, low-contrast, and weakly salient. These factors create a clear gap between generic pre-trained representations and the discriminative cues required for robust defect detection~\cite{baeuerle2025foundation}.

To bridge this gap, we propose \textit{BladeYOLO}, a YOLOv12-L-based defect detector with a DINOv3-initialized ViT-S backbone. In this way, the model preserves the generalization capability of large-scale self-supervised pre-training while adapting to the visual characteristics of blade defect inspection. The overall architecture is illustrated in \autoref{fig:fig_2}.

The design of BladeYOLO is guided by three complementary considerations. First, to alleviate the limitations caused by scarce training data, we employ a DINOv3-initialized ViT-S backbone to transfer large-scale self-supervised representation priors to wind turbine blade defect detection. Second, to improve the perception of small and weak-saliency defects, we develop an \emph{MWE} module, which consists of the Detail-Enhanced Multi-scale Branch for preserving high-frequency structural cues and a Cross-Mamba module for progressively propagating high-level semantic guidance to shallow features. In this way, subtle defect details are first enhanced and then semantically calibrated in a coarse-to-fine manner. Third, to improve robustness against environmental appearance variations caused by illumination changes and adverse atmospheric conditions, we introduce a lightweight \emph{Style-Injector} module. This module extracts environment-related style cues via frequency-domain decomposition and injects them into selected self-attention blocks, enabling more stable feature representation under diverse inspection conditions.

Through the coordinated integration of these components, BladeYOLO adapts the DINOv3-initialized ViT-S backbone to wind turbine blade defect detection from three aspects: transferable representation under data-scarce conditions, robustness to environmental variations, and enhanced perception of weak-saliency defects. As a result, the proposed framework achieves more reliable multi-scale feature modeling and more accurate defect detection in complex real-world inspection scenarios.

\subsection{DINOv3-initialized ViT Backbone}

To alleviate the performance degradation caused by scarce training data, we employ a Vision Transformer initialized with DINOv3 pre-trained weights as the semantic backbone of BladeYOLO. Compared with conventional CNN backbones trained only on task-specific data, the DINOv3-initialized ViT provides stronger transferable representation priors learned from large-scale self-supervised pre-training, which is particularly beneficial for wind turbine blade defect detection under limited-data conditions.

Specifically, we replace the original backbone of YOLOv12-L with a ViT-S model initialized with DINOv3 pre-trained weights. Given an input image $x \in \mathbb{R}^{3 \times H \times W}$, the ViT backbone first divides the image into a sequence of fixed-size patches and projects them into token embeddings, which are then processed by stacked self-attention blocks to model long-range dependencies and global semantic context. Since a plain ViT backbone typically produces features at a single native resolution, we extract intermediate semantic features from \rev{the 4th, 8th, and 12th transformer blocks} and project them to three scales, denoted by $\{D_3, D_4, D_5\}$, corresponding to spatial resolutions of $\frac{1}{8}$, $\frac{1}{16}$, and $\frac{1}{32}$ of the input image, respectively. These features serve as the semantic branch of the proposed framework.

The extracted semantic features are subsequently fused with the outputs of the Detail-Enhanced Multi-scale Branch at the corresponding scales, forming the input feature pyramid for the Cross-Mamba module. In this way, the DINOv3-initialized ViT backbone provides strong semantic representations, while the subsequent MWE module further complements it with fine-grained structural details and progressive semantic calibration.

\subsection{Mamba-guided Weak-Defect Enhancement}

Although the DINOv3-initialized ViT backbone provides strong global semantic representations, its patch-based tokenization and single-resolution feature output make it less sensitive to subtle local structures. This limitation is particularly unfavorable for wind turbine blade defects, which are often small-scale and weakly salient. To compensate for this weakness, we develop an MWE module, which consists of a Detail-Enhanced Multi-scale Branch and a Cross-Mamba module. \autoref{fig:fig_2} (a) shows the overall architecture.

\subsubsection{Detail-Enhanced Multi-scale Branch}

We first apply a stem consisting of a $3\times3$ convolution followed by max pooling to obtain a base feature map $X \in \mathbb{R}^{C \times \frac{H}{4} \times \frac{W}{4}}$. The stem feature is then fed into three parallel branches to generate detail-enhanced features at three resolutions:
$
\{L_{\mathrm{fine}}, L_{\mathrm{medium}}, L_{\mathrm{large}}\}$,
where
$L_{\mathrm{fine}} \in \mathbb{R}^{C_f \times \frac{H}{8} \times \frac{W}{8}}$,
$L_{\mathrm{medium}} \in \mathbb{R}^{C_m \times \frac{H}{16} \times \frac{W}{16}}$, and
$L_{\mathrm{large}} \in \mathbb{R}^{C_l \times \frac{H}{32} \times \frac{W}{32}}$.

The \emph{Fine} branch is designed to preserve local structural details for tiny defects. It applies two standard convolutions and a 1-level wavelet convolution to enhance high-frequency components:
\begin{equation}
L_{\mathrm{fine}} = \operatorname{WTConv}_1\!\big(\operatorname{Conv}(\operatorname{Conv}(X))\big).
\end{equation}
Here, $\operatorname{WTConv}_1(\cdot)$ performs 1-level wavelet decomposition, subband-wise enhancement, and inverse wavelet reconstruction~\cite{finder2024wavelet}. This branch focuses on retaining subtle edges and fine structural discontinuities that are easily weakened in patch-based representations.

The \emph{Medium} branch is used to capture intermediate-scale structures that are beneficial for moderate-size defects. Starting from $X$, it applies a sequence of convolutions to produce a feature map at $\frac{1}{16}$ resolution:
$
L_{\mathrm{medium}} = \operatorname{Conv}_{m}(X)
$, where $\operatorname{Conv}_{m}(\cdot)$ denotes the convolutional stack.

The \emph{Large} branch is introduced to enlarge the receptive field and provide coarse yet semantically informative context. It first applies a 2-level wavelet convolution to $X$, and then downsamples the output to $\frac{1}{32}$ resolution using two stride-2 convolutions:
\begin{equation}
L_{\mathrm{large}} = \operatorname{Conv}_{s=2}\!\big(\operatorname{Conv}_{s=2}(\operatorname{WTConv}_2(X))\big).
\end{equation}
\rev{Here, $\operatorname{WTConv}_2(\cdot)$ denotes a 2-level wavelet convolution. Compared with the 1-level version, it further decomposes the low-frequency component to enlarge the effective receptive field while retaining high-frequency structural information. In our implementation, the fine branch uses $\operatorname{WTConv}_1(\cdot)$ for local detail enhancement, while the large branch uses $\operatorname{WTConv}_2(\cdot)$ for coarse contextual modeling.}

The resulting detail-enhanced features at three resolutions are denoted by $\{L_3, L_4, L_5\}$, corresponding to the fine, medium, and large branches, respectively. These features are concatenated with the semantic features extracted from the DINOv3-initialized ViT at the corresponding scales to form a fused feature pyramid, which is subsequently refined by the proposed Cross-Mamba module before being fed into the YOLOv12-L neck and detection head.

\subsubsection{Cross-Mamba Module}

Although the combination of the DINOv3-initialized ViT and the Detail-Enhanced Multi-scale Branch provides both global semantics and local detail cues, the shallow high-resolution features may still suffer from semantic ambiguity when defects are subtle, low-contrast, or embedded in cluttered blade textures. In such cases, shallow features preserve rich spatial details but often lack sufficiently stable semantic guidance, which can lead to false positives or missed detections. Moreover, simple feature fusion mainly performs static information aggregation and is insufficient to explicitly resolve the semantic ambiguity of shallow responses. To address this issue, we introduce a Cross-Mamba module that explicitly propagates high-level semantic priors to shallow feature maps through the linear-time sequence modeling capability of Mamba~\cite{zhu2024vision}. \rev{Intuitively, this module acts as a cross-scale semantic guidance mechanism that strengthens shallow detail features with deeper semantic cues.}

\begin{figure}[t!]
    \centering
    \includegraphics[width=1.0\columnwidth]{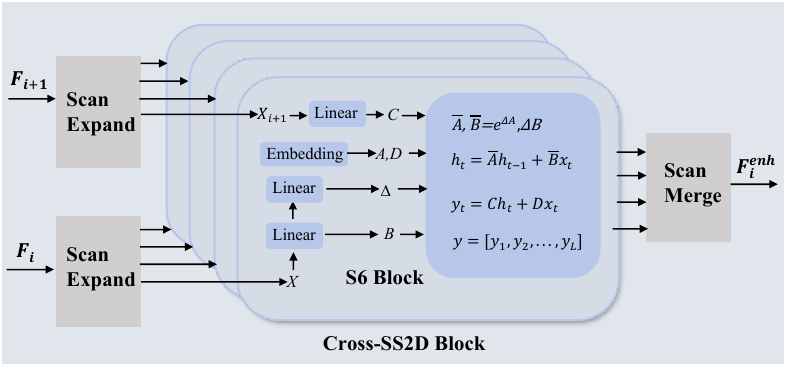}
    \caption{Architecture of Cross-SS2D block. The target feature $F_i$ and the adjacent higher-level guidance feature $F_{i+1}$ are first expanded into four scanning sequences. Each sequence is then processed by an S6 block, where the target feature determines the state-transition dynamics and the guidance feature modulates the output projection. The resulting sequences are finally merged to produce the semantically calibrated feature $F_i^{\mathrm{enh}}$.}
    \label{fig:fig_5}
\end{figure}

The core component of the Cross-Mamba module is a Cross-SS2D block, as shown in \autoref{fig:fig_5}. Different from conventional self-attention, which models dependencies within a single feature scale, Cross-SS2D is designed for cross-scale semantic guidance. 
\rev{For a target feature $F_i$ and its adjacent higher-level feature $F_{i+1}$, the higher-level feature is first upsampled to the spatial resolution of $F_i$ and denoted as $G_{i+1}$. The target feature $F_i$ provides detailed spatial structures, while the aligned guidance feature $G_{i+1}$ provides stronger semantic information. In the Cross-SS2D block, the target feature is used to determine the state-transition dynamics, and the guidance feature modulates the output projection. In this way, the module preserves the spatial details of shallow features while introducing high-level semantic guidance.}

Specifically, the cross-scale guidance process of the proposed Cross-Mamba module is shown in \autoref{fig:fig_2} (b). After fusing the multi-scale ViT features and $\{L_3, L_4, L_5\}$, we obtain a feature pyramid
$\mathcal{F} = \{F_3, F_4, F_5\}$, where $F_5 \in \mathbb{R}^{C_5 \times \frac{H}{32} \times \frac{W}{32}}$ is the deepest semantic feature, $F_4 \in \mathbb{R}^{C_4 \times \frac{H}{16} \times \frac{W}{16}}$ is the intermediate feature, and $F_3 \in \mathbb{R}^{C_3 \times \frac{H}{8} \times \frac{W}{8}}$ is the shallow high-resolution feature. We adopt a progressive two-stage semantic enhancement strategy. First, $F_5$ is used to guide $F_4$:
\begin{equation}
F_4^{\mathrm{enh}} = \operatorname{CrossMamba}(F_4, F_5).
\end{equation}
Then, the enhanced intermediate feature $F_4^{\mathrm{enh}}$ is further used to calibrate the shallow feature:
\begin{equation}
F_3^{\mathrm{enh}} = \operatorname{CrossMamba}(F_3, F_4^{\mathrm{enh}}).
\end{equation}
Here, $\operatorname{CrossMamba}(\cdot,\cdot)$ denotes the cross-scale semantic calibration operator implemented by Cross-SS2D. This progressive coarse-to-fine design avoids directly injecting deep semantics into shallow features across a large resolution gap, thereby making semantic transfer more stable and effective.

Finally, the enhanced feature set $\{F_3^{\mathrm{enh}}, F_4^{\mathrm{enh}}, F_5\}$ is fed into the YOLOv12-L neck and detection head for defect classification and localization.

\subsection{Style Injection}

To improve the robustness of the DINOv3 backbone against environment-induced appearance shifts in wind turbine blade inspection, we propose a \emph{Style-Injector} module, as shown in \autoref{fig:fig_2}(c), to extract style cues from the Fourier amplitude spectrum and inject them into selected ViT attention blocks via a cross-attention adapter.

\begin{figure}[tbp]
  \centering
  \includegraphics[width=1\linewidth]{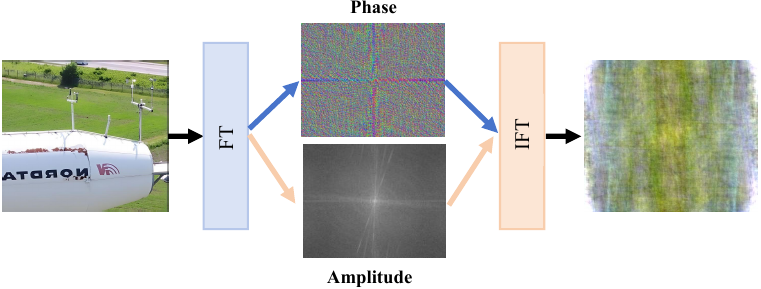}
  \caption{Illustration of style extraction in the proposed Style-Injector. For an input wind turbine blade image, the Fourier transform (FT) is first applied to obtain its phase and amplitude spectra. A style image is then reconstructed through inverse Fourier transform (IFT) using the amplitude spectrum together with an averaged phase, which suppresses fine-grained structural details while preserving global appearance statistics.}
  \label{fig:fig_3}
\end{figure}

In real inspection scenarios, blade images often exhibit global shifts in color, contrast, and visibility due to illumination changes and haze. \rev{The Fourier amplitude spectrum mainly reflects global appearance statistics, such as illumination, contrast, and color distribution, whereas the phase spectrum is more related to spatial structure and fine-grained content~\cite{lyu2025efficient}.}
\rev{Therefore, we use the amplitude spectrum to construct a style image $\hat{x}$ that preserves environment-related appearance information while suppressing fine structural details, as illustrated in \autoref{fig:fig_3}.} 
The reconstructed image $\hat{x}$ is then fed into a lightweight convolutional encoder $E_s$ followed by global average pooling (GAP) to obtain a compact style descriptor:
\begin{equation}
p_{\text{style}} = \operatorname{GAP}\!\left(E_s(\hat{x})\right).
\end{equation}

To inject these style cues without disrupting the semantic priors of the pre-trained ViT, we adopt a parameter-efficient adapter design~\cite{houlsby2019parameter}. Let $Z \in \mathbb{R}^{N \times d}$ denote the input token sequence of a ViT attention block. The style descriptor $p_{\text{style}}$ is first projected into a style token $s$ through an MLP, and a cross-attention branch is then used to enhance the frozen self-attention output:
\begin{equation}
\tilde{Z} = \operatorname{MHA}(Z) + \operatorname{CrossAttn}(Q=Z, K=s, V=s),
\end{equation}
\rev{where $\operatorname{MHA}(\cdot)$ denotes the original multi-head self-attention operation, and $\operatorname{CrossAttn}(\cdot)$ injects the style token into the image token sequence. The self-attention parameters inherited from DINOv3 are kept frozen, while the style projection layer and cross-attention branch are trainable.}

In addition, a bottleneck adapter is introduced alongside the frozen FFN:
\begin{equation}
Z_{\text{out}} = \operatorname{FFN}(\tilde{Z}) + W_u\,\sigma\!\left(W_d \tilde{Z}\right),
\end{equation}
where $W_d$ and $W_u$ are learnable projections for dimensionality reduction and restoration, respectively, and $\sigma(\cdot)$ denotes the GELU activation function. 
\rev{During training, the pre-trained FFN parameters remain frozen, and only the newly introduced adapter parameters are optimized.}

To balance adaptation capability and computational overhead, we insert the Style-Injector only into a subset of ViT attention blocks, as shown in \autoref{fig:fig_2} (c). 
\rev{Specifically, we insert the Style-Injector into the 4th, 8th, and 12th ViT blocks. This sparse insertion strategy allows the DINOv3-initialized backbone to adapt to environment-induced appearance variations while preserving the semantic priors learned from large-scale self-supervised pre-training.}

\section{Results and Discussion}
\label{sec:experiments}
\subsection{Wind Turbine Blade Defect Datasets}

\textbf{WTBlade-Defect dataset}. 
In this study, we constructed the WTBlade-Defect dataset using 1,785 images of wind turbine blade defects from publicly available online sources. The original image resolutions range from $586 \times 371$ to $640 \times 640$ pixels. Before being fed into the model, all images were automatically padded to a unified resolution of $640 \times 640$ pixels. Then, to improve data diversity and alleviate the limitation of insufficient training samples, several data augmentation strategies were applied, including horizontal and vertical flipping with a probability of $50\%$, the addition of Gaussian noise with an intensity of $10\%$, random sharpening within $20\%$, and random brightness adjustment within $20\%$. After augmentation, the dataset was expanded to 5,326 images in total. 

\begin{figure}[]
  \centering
  \includegraphics[width=1.0\linewidth]{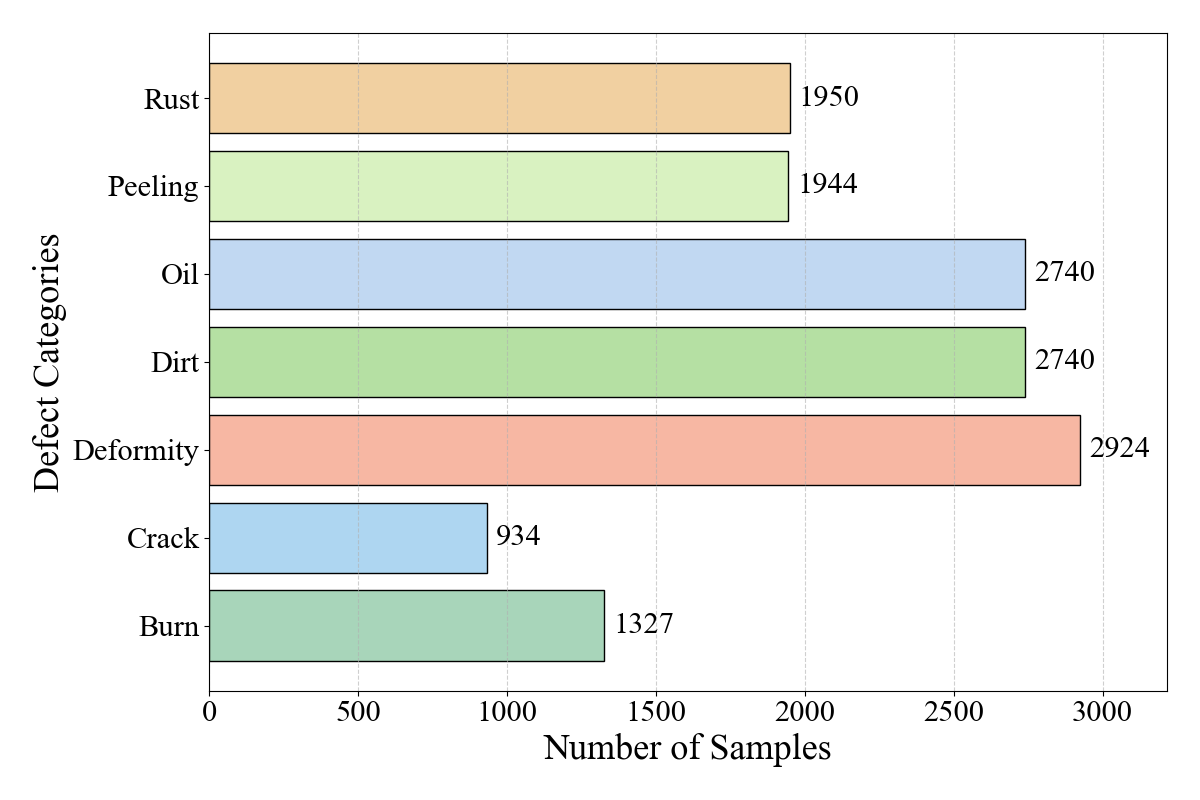}
  \caption{Category distribution of defect instances in the WTBlade-Defect dataset. The dataset contains seven defect categories, namely burn, crack, deformity, dirt, oil, peeling, and rust, with the corresponding number of annotated instances shown for each category.}
  \label{fig:fig_wind}
\end{figure}

\begin{figure}[]
  \centering
  \includegraphics[width=1.0\linewidth]{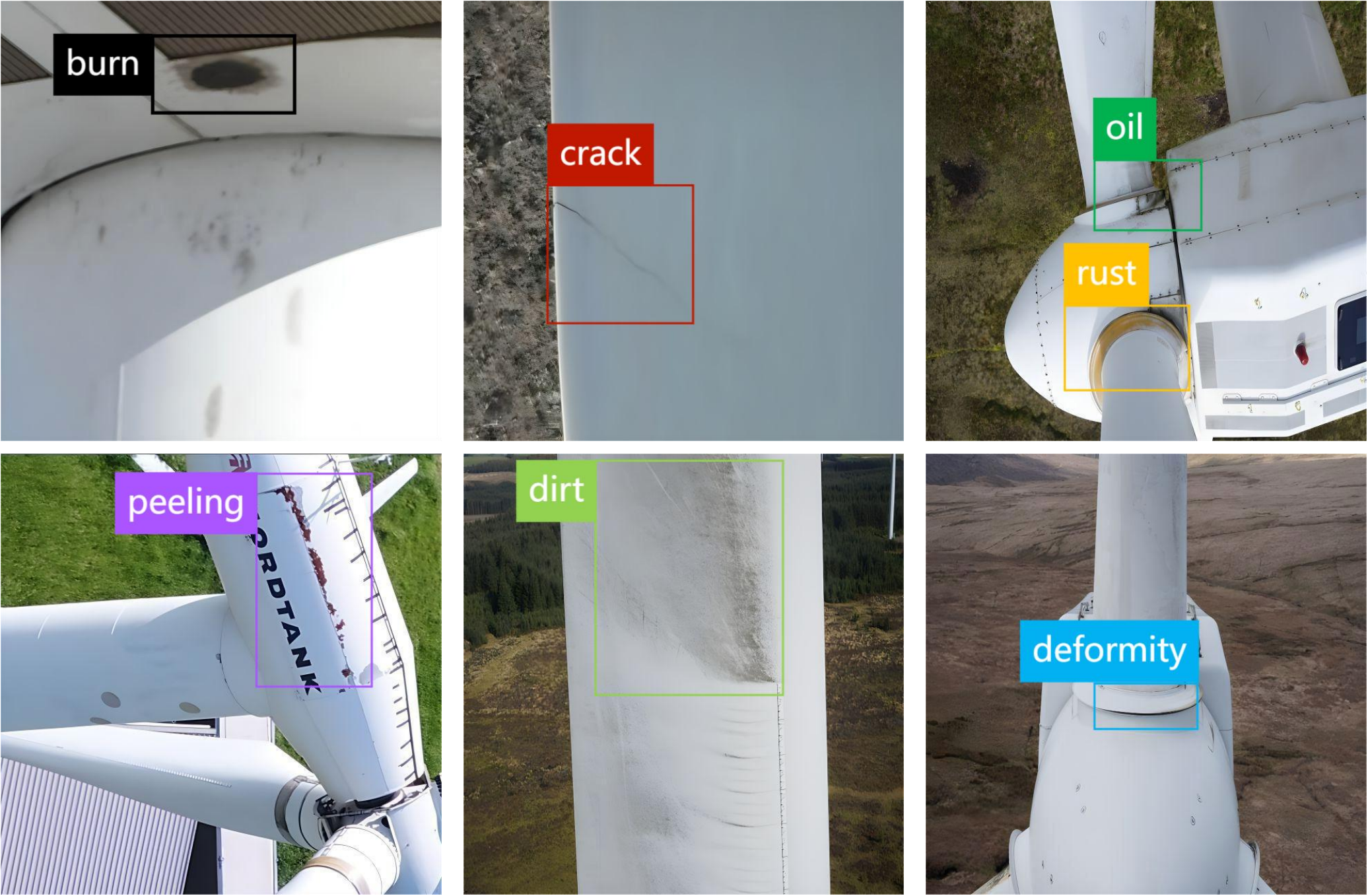}
  \caption{Representative samples from the WTBlade-Defect dataset. The examples cover seven typical blade-surface defect categories, including burn, crack, deformity, dirt, oil, peeling, and rust, illustrating the diversity of defect appearance, scale, and background complexity in real inspection scenarios.}
  \label{fig:fig_exp}
\end{figure}

The resulting WTBlade-Defect dataset contains seven defect categories, including 1,327 \emph{burn} instances, 934 \emph{crack} instances, 2,924 \emph{deformity} instances, 2,740 \emph{dirt} instances, 2,740 \emph{oil} instances, 1,944 \emph{peeling} instances, and 1,950 \emph{rust} instances. The category distribution is illustrated in \autoref{fig:fig_wind}. Following a $9{:}1$ split, 4,766 images were used for training and 560 images were used for testing. Example images from the \emph{WTBlade-Defect} dataset are shown in \autoref{fig:fig_exp}.

\textbf{Public Wind Surface Defect dataset}. 
To further evaluate the robustness of \emph{BladeYOLO} on a related public dataset, we conducted experiments on the Wind Surface Defect dataset introduced by Liu et al.~\cite{Liu2025Wind}. This dataset contains 3,790 wind turbine blade images with five defect categories, including 2,116 \emph{corrosion} instances, 2,344 \emph{hide-craze} instances, 412 \emph{surface-eye} instances, 480 \emph{thunderstrike} instances, and 1,162 \emph{dirt} instances. Following the experimental protocol in~\cite{Liu2025Wind}, the dataset was randomly divided into training and test sets with a ratio of $4{:}1$.

\subsection{Experimental Environment and Evaluation Metrics}

All experiments were conducted on a workstation running Windows 10, equipped with an Intel(R) Core(TM) i7-11700 CPU and a single NVIDIA RTX 4090 GPU with 24 GB memory. The input image size was set to $640 \times 640$ for all compared methods. Stochastic gradient descent was adopted as the optimizer, with an initial learning rate of 0.01 and a cosine annealing learning rate schedule. The batch size was set to 10 to avoid out-of-memory issues during training. \rev{Except for the ViT backbone initialized with DINOv3 pre-trained weights, the newly introduced MWE module, Style-Injector, YOLOv12-L neck, and detection head were randomly initialized and trained from scratch under the same training settings.} Under these experimental settings, the proposed model required 16.9 h to complete 300 training epochs.

All involved methods were evaluated from two perspectives: detection performance and computational efficiency. Detection performance was measured using Average Precision (AP) and mean Average Precision (mAP). Computational efficiency was characterized by the number of parameters (Params) and the number of floating-point operations (FLOPs).

\begin{figure*}[]
  \centering
\includegraphics[width=0.85\textwidth]{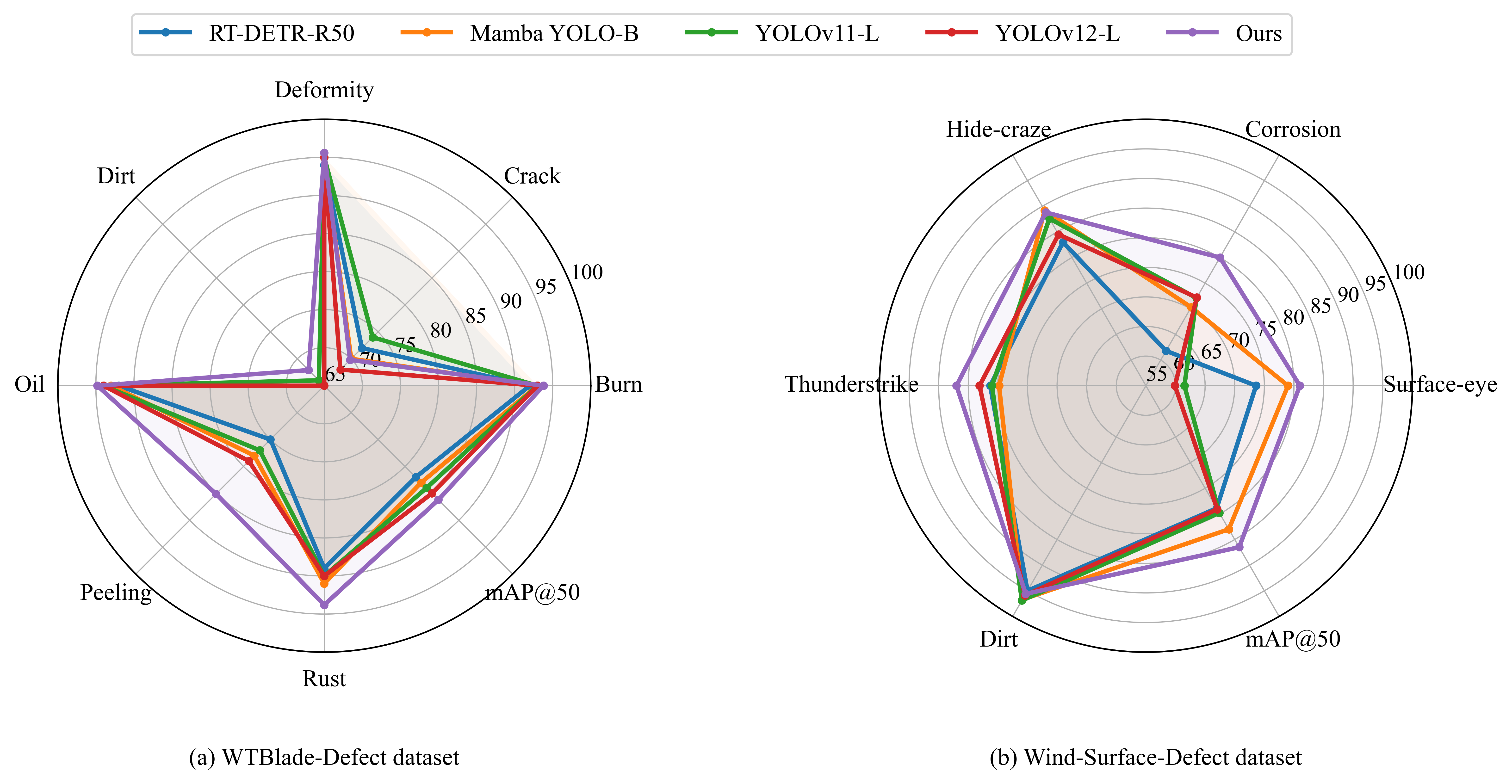}
  \caption{Radar-chart comparison of several representative detection methods on the (a) WTBlade-Defect dataset and (b) Wind Surface Defect dataset, in terms of category-wise AP$_{50}$ and overall mAP$_{50}$. The axes corresponding to defect categories denote the AP$_{50}$ values of different methods for the respective defect classes. In contrast, the mAP$_{50}$ axis indicates the overall mean detection accuracy of each method on the entire dataset.}
  \label{fig:radar_sci}
\end{figure*}

\begin{figure*}[]
  \centering
  \includegraphics[width=1.0\textwidth]{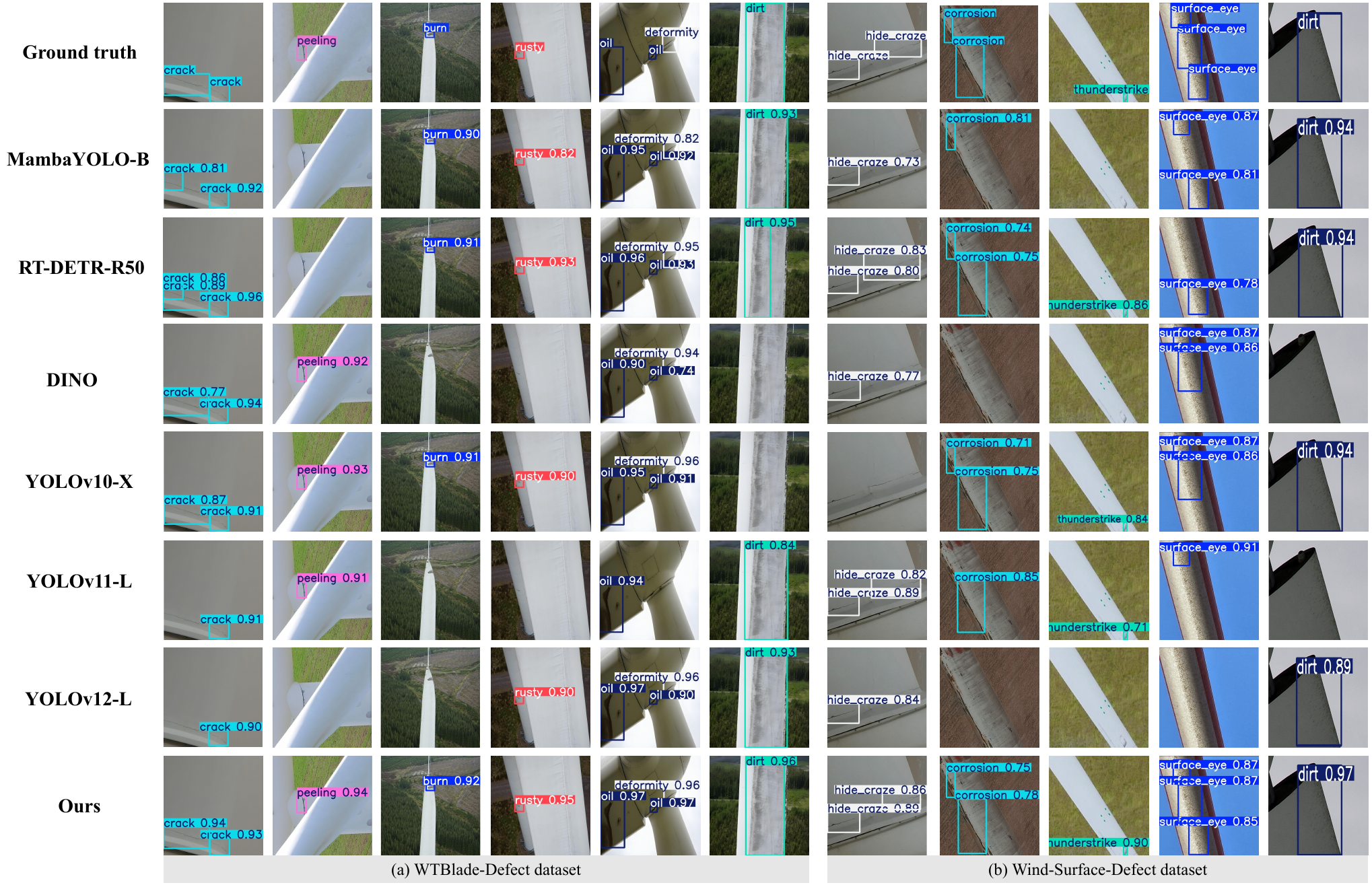}
  \caption{Qualitative comparison of detection results produced by different methods on two wind turbine blade defect datasets: (a) WTBlade-Defect dataset and (b) Wind Surface Defect dataset. For each example, the predicted bounding boxes are annotated with category labels and confidence scores. Compared with the competing methods, the proposed method yields more accurate localization and more reliable classification, especially in challenging cases involving weak-saliency defects, small targets, and complex backgrounds.}
  \label{fig:fig_sample}
\end{figure*}

\begin{figure*}[]
    \centering
\includegraphics[width=\textwidth]{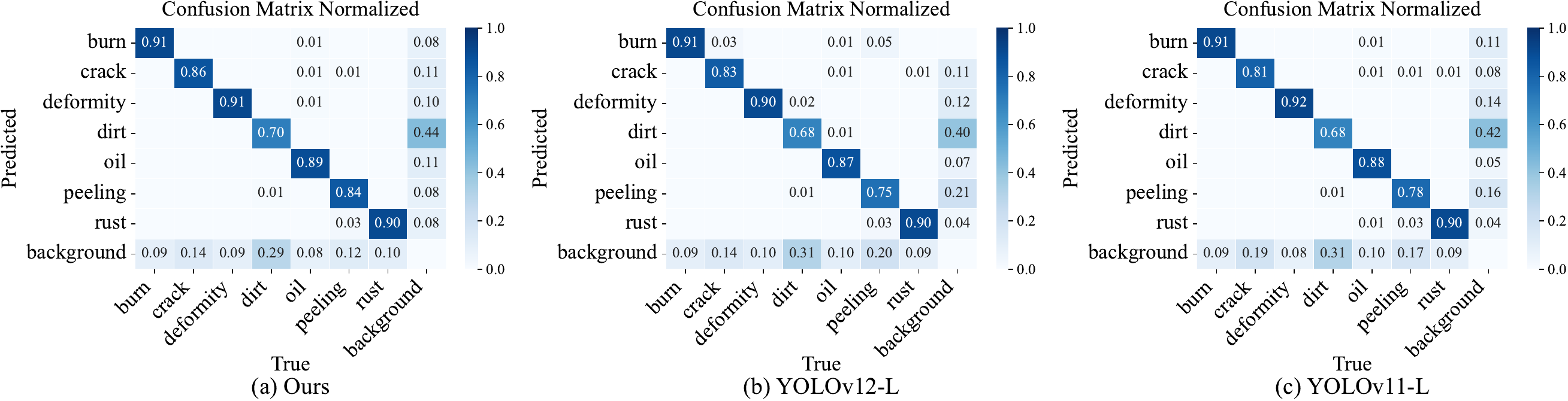}
    \caption{Normalized confusion matrices of different models on the WTBlade-Defect dataset: (a) Ours; (b) the baseline YOLOv12-L; (c) YOLOv11-L. Compared to the baseline models, our proposed method demonstrates superior classification accuracy across most defect categories. Notably, our model effectively reduces the misclassification rate of dirt and peeling defects, achieving  more balanced and robust performance in complex industrial environments.}
    \label{fig:fig_metric}
\end{figure*}

\subsection{Comparative Experiments}

To comprehensively evaluate the effectiveness of the proposed \emph{BladeYOLO} for wind turbine blade defect detection, we compare it with a wide range of representative object detection methods, including transformer-based detectors (the RT-DETR series)~\cite{zhao2024detrs,lv2024rt}, recent YOLO-series models~\cite{wang2024yolov9,wang2024yolov10,khanam2024yolov11}, and MambaYOLO-series models~\cite{wang2025mamba}. 
\rev{
To ensure a fair and transparent comparison, all methods are evaluated on the same dataset split and test set, with the same input resolution and evaluation metrics, while following the standard inference settings of their respective implementations. For each compared model, we follow commonly used or officially recommended training settings and select the best-performing setting on our detection task when multiple standard options are available, to avoid disadvantaging any baseline. In addition, \emph{BladeYOLO} does not use additional training data, task-specific data augmentation, or method-specific post-processing.
}

\subsubsection{Quantitative Comparison}
The quantitative results of all compared methods are reported in \autoref{tab:compare}.
\rev{
The mAP$_{50}$ and mAP$_{50-95}$ results are reported as the mean and standard deviation over three random seeds (0, 1, and 2), rather than as single-run values. This repeated-run evaluation provides a more reliable assessment of performance robustness across different random initializations.
}
The proposed method achieves superior overall detection performance compared with the competing methods. Specifically, \emph{BladeYOLO} obtains an mAP$_{50}$ of 85.7$\pm$0.1\% and an mAP$_{50-95}$ of 74.7$\pm$0.1\%, demonstrating clear advantages over several strong baselines. These results verify that the proposed architecture can effectively improve wind turbine blade defect detection under challenging inspection conditions.


To provide a more intuitive view of category-level performance, we further visualize the results of several representative methods, including RT-DETR-R50, MambaYOLO-B, YOLOv11-L, YOLOv12-L, and the proposed method, using a radar chart. As shown in \autoref{fig:radar_sci} (a), the proposed method forms the largest enclosed area, indicating a more balanced and consistently strong detection capability across different defect categories. Notably, its advantages are more evident on challenging categories such as \emph{peeling}, \emph{rust}, and \emph{deformity}, which further validates its robustness.

Overall, the above quantitative results demonstrate that \emph{BladeYOLO} achieves a favorable balance between detection accuracy and model complexity, making it a competitive solution for practical wind turbine blade defect detection.

\subsubsection{Qualitative Comparison}
Representative detection examples are shown in \autoref{fig:fig_sample}. As can be observed, \emph{BladeYOLO} is able to more accurately localize and classify a variety of blade-surface defects, even in challenging scenarios involving weak-saliency targets, small defects, and cluttered backgrounds. Compared with the competing methods, the proposed method produces more reliable predictions with fewer missed detections and false alarms. These visual results further demonstrate the practical effectiveness and real-world applicability of \emph{BladeYOLO} for wind turbine blade defect detection.

To further investigate the error patterns and inter-class confusion of different methods, we present the normalized confusion matrices of several representative models in \autoref{fig:fig_metric}. Compared with the baseline models YOLOv12-L and YOLOv11-L, the proposed method exhibits a clearer diagonal structure, indicating better category separability. More importantly, the confusion with the background is noticeably reduced, especially for challenging categories such as \emph{dirt} and \emph{peeling}. This observation suggests that the proposed modules can enhance defect awareness while effectively suppressing background interference under complex background conditions.

\subsubsection{Weak-saliency Subset Comparison}
\rev{
To further examine the detection performance of representative detectors under weak-saliency conditions, we construct a weak-saliency subset from the WTBlade-Defect test set. It should be noted that weak saliency is not treated as an additional defect category, but as an instance-level attribute for constructing a challenging evaluation subset. The original defect categories remain unchanged during evaluation.
}

\rev{
Previous saliency studies have shown that salient targets become more difficult to perceive when they are weakly distinguishable from the surrounding background~\cite{tong2015salient,mu2019salient,fan2018salient}. Inspired by these studies and the small-defect characteristics of blade inspection, we define weak-saliency defects in this work using two measurable instance-level properties: relative area and local contrast.
}

\rev{
For each ground-truth defect instance, the relative area is defined as $
A_r = \frac{w \times h}{W \times H},
$
where \(w\) and \(h\) denote the width and height of the ground-truth bounding box, and \(W\) and \(H\) denote the width and height of the image. The local contrast is computed as
$
C_l = |\mu_{\mathrm{in}} - \mu_{\mathrm{out}}|,
$
where \(\mu_{\mathrm{in}}\) is the mean grayscale intensity inside the bounding box, and \(\mu_{\mathrm{out}}\) is the mean grayscale intensity of its surrounding region. The surrounding region is obtained by enlarging the ground-truth bounding box by 1.5 times around its center and removing the original bounding-box region, which provides an approximate local background for estimating contrast.
}

\rev{
In this study, weak-saliency defects are defined as instances whose \(A_r\) and \(C_l\) both fall within the lowest 30\% of their respective test-set distributions, corresponding to defects that are simultaneously small in size and low in local contrast. This percentile-based criterion avoids manually selecting dataset-specific absolute thresholds while retaining sufficiently challenging samples for reliable evaluation.
}

\rev{
Based on the identified weak-saliency instances, we construct a weak-saliency image subset for evaluation. An image is selected if it contains at least one weak-saliency defect. Following this rule, 107 weak-saliency defect instances are identified from 1,017 ground-truth instances in the WTBlade-Defect test set, resulting in 81 selected images. The original labels and defect categories of these images are retained for standard detection evaluation.
}

\rev{
Note that this subset is exclusively used for focused evaluation and is not involved in training or hyperparameter selection.
For comparison, we select representative methods from different detector families, including RT-DETR-R50, RTMDet-L, MambaYOLO-B, YOLOv8-L as a representative YOLO baseline, YOLOv12-L as the direct baseline, and the proposed BladeYOLO.
}

\rev{
As shown in \autoref{tab:weak_subset_comparison}, the weak-saliency subset provides a focused evaluation of detection performance under small-scale and low-contrast defect conditions. BladeYOLO achieves the best performance, with 84.1\% mAP$_{50}$ and 73.7\% mAP$_{50-95}$. Compared with the direct YOLOv12-L baseline, BladeYOLO improves mAP$_{50}$ and mAP$_{50-95}$ by 2.6 and 3.3 percentage points, respectively. It also outperforms YOLOv8-L and MambaYOLO-B, indicating that the proposed weak-defect enhancement design is effective beyond the general improvements brought by recent YOLO and Mamba-based detectors. These results provide focused quantitative evidence that BladeYOLO achieves more favorable performance under weak-saliency conditions, in addition to improving overall detection performance on the full test set.
}

\begin{table*}[h!]
\small
\centering
\caption{Quantitative comparison of different defect detection methods on the WTBlade-Defect dataset. Category-wise AP$_{50}$, mAP$_{50}$, mAP$_{50-95}$, model parameters, and FLOPs are reported. The mAP$_{50}$ and mAP$_{50-95}$ results are reported as mean $\pm$ standard deviation over three random seeds (0, 1, and 2). The top three results for each metric are highlighted in \TopOne{red}, \TopTwo{blue}, and \TopThree{green}, respectively.}
\label{tab:compare}
\resizebox{\textwidth}{!}{
  \begin{tabular}{lccccccccccc}
  \toprule
  \multirow{2}{*}{Method} 
  & \multicolumn{7}{c}{${\rm AP}_{50}$ (\%)} 
  & \multirow{2}{*}{mAP$_{50}$ (\%, mean $\pm$ std)} 
  & \multirow{2}{*}{mAP$_{50-95}$ (\%, mean $\pm$ std)} 
  & \multirow{2}{*}{Params} 
  & \multirow{2}{*}{FLOPs (G)} \\
  \cmidrule(lr){2-8}
   & peeling & rust & deformity & dirt & oil & burn & crack & & & & \\
  \midrule
  Tood-R50        & 76.5 & 88.4 & 91.9 & 66.2 & 91.5 & 91.2 & 57.3 & 80.4$\pm$0.1 & 55.5$\pm$0.2 & 39.2M  & 80.4 \\
  DINO-4Scale-R50 & 81.5 & 86.9 & 93.3 & 65.0 & 94.2 & \TopOne{94.4} & 59.8 & 82.2$\pm$0.2 & 53.3$\pm$0.1 & 47.2M  & 119 \\
  RT-DETR-R50     & 74.8 & 88.4 & 90.1 & 62.2 & 90.2 & 92.8 & 67.2 & 80.8$\pm$0.1 & 62.6$\pm$0.1 & 41.6M & 80.4 \\
  RTMDet-L        & 79.7 & 80.1 & 91.1 & 51.9 & 74.6 & 90.1 & 58.6 & 79.9$\pm$0.2 & 56.5$\pm$0.1 & 24.7M & 80.23 \\
  MambaYOLO-B    & \TopTwo{84.7} & 92.5 & \TopThree{95.2} & 67.7 & 94.4 & \TopTwo{94.2} & 67.7 & \TopTwo{85.2$\pm$0.1} & 72.9$\pm$0.1 & 21.8M & 49.6 \\
  MambaYOLO-L    & \TopThree{83.8} & 92.5 & 94.4 & 62.1 & 93.2 & \TopThree{94.0} & 65.8 & 83.7$\pm$0.2 & 69.5$\pm$0.1 & 57.6M & 156 \\
  YOLOv8-L        & 82.5 & 91.3 & 94.2 & \TopOne{68.9} & 93.9 & 92.9 & \TopOne{71.8} & \TopThree{84.9$\pm$0.1} & \TopTwo{74.2$\pm$0.2} & 43.6M & 164.8 \\
  YOLOv9-C        & 81.3 & \TopTwo{93.0} & \TopOne{96.0} & 67.5 & \TopThree{94.7} & 92.4 & 69.4 & 84.9$\pm$0.2 & \TopThree{74.2$\pm$0.1} & 25.3M & 102.3 \\
  YOLOv10-X       & 80.4 & 89.9 & 93.3 & 64.0 & 92.4 & 92.2 & 66.7 & 82.7$\pm$0.1 & 71.7$\pm$0.2 & 31.6M & 169.8 \\
  YOLOv11-L       & 79.5 & 91.9 & 93.8 & \TopTwo{68.1} & \TopOne{95.4} & 93.5 & \TopTwo{70.1} & 84.8$\pm$0.2 & 73.9$\pm$0.2 & 25.3M & 86.6 \\
  YOLOv12-X       & 82.3 & \TopThree{92.5} & 94.4 & 66.4 & 93.7 & 93.4 & 63.9 & 83.8$\pm$0.2 & 73.3$\pm$0.2 & 59.2M & 184.1 \\
  BladeYOLO (Ours) & \TopOne{85.1} & \TopOne{93.8} & \TopTwo{95.6} & \TopThree{67.7} & \TopTwo{94.8} & 93.8 & \TopThree{69.6} & \TopOne{85.7$\pm$0.1} & \TopOne{74.7$\pm$0.1} & 43.8M & 167.7 \\
  \bottomrule
  \end{tabular}
}
\end{table*}

\begin{table}[t]
\centering
\caption{\rev{Performance comparison of representative detectors on the weak-saliency subset of WTBlade-Defect.}}
\label{tab:weak_subset_comparison}
\resizebox{\columnwidth}{!}{
\begin{tabular}{lcc}
\hline
Method & mAP$_{50}$ (\%) & mAP$_{50-95}$ (\%) \\
\hline
RT-DETR-R50 & 80.2 & 61.5 \\
RTMDet-L & 77.2 & 53.2 \\
MambaYOLO-B & 82.3 & 71.9 \\
YOLOv8-L & 81.8 & 72.3 \\
YOLOv12-L & 81.5 & 70.4 \\
BladeYOLO (Ours) & 84.1 & 73.7 \\
\hline
\end{tabular}
}
\vspace{-2mm}
\end{table}

\begin{table*}[t]
\centering
\scriptsize
\setlength{\tabcolsep}{2pt}
\caption{Ablation study of the proposed method on the WTBlade-Defect test set. The first row reports the baseline YOLOv12-L model, while the remaining rows present ablation variants based on the DINOv3-initialized backbone with different combinations of the proposed modules. AP$_{50}$ denotes the average precision at an IoU threshold of 0.5 for each defect category, whereas mAP$_{50}$ and mAP$_{50-95}$ represent the mean AP over all categories at IoU thresholds of 0.5 and 0.5:0.95, respectively. A checkmark ($\checkmark$) indicates that the corresponding module is enabled. The best results among the DINOv3-based ablation variants are highlighted in bold.}
\label{tab:module_comparison}

\resizebox{\textwidth}{!}{
\begin{tabular}{lccccccccccccc}
\toprule
\multirow{2}{*}{Backbone} &
\multirow{2}{*}{Style-Injector} &
\multirow{2}{*}{Detail-Enhanced Multi-scale Branch} &
\multirow{2}{*}{Cross-Mamba} &
\multicolumn{7}{c}{AP$_{50}$ (\%)} &
\multirow{2}{*}{mAP$_{50}$ (\%)} &
\multirow{2}{*}{mAP$_{50-95}$ (\%)} \\
\cmidrule{5-11}
& & & & burn & crack & deformity & dirt & oil & peeling & rust & & \\
\midrule
YOLOv12-L &  &  &  &
94.6 & 65.7 & 94.1 & 65.0 & 92.3 & 78.8 & 92.7 & 83.3 & 71.9 \\

DINOv3 (baseline) &  &  &  &
90.9 & 65.7 & 94.5 & 66.6 & 95.1 & 84.2 & 92.4 & 84.2 & 66.6 \\

DINOv3 & $\checkmark$ &  &  &
93.1 & 68.3 & 95.6 & 67.3 & 95.2 & 84.6 & 92.8 & 85.3 & 68.2 \\

DINOv3 & $\checkmark$ & $\checkmark$ &  &
93.7 & 68.6 & \textbf{95.8} & 67.4 & \textbf{95.8} & 84.6 & 92.5 & 85.7 & 71.1 \\

DINOv3 & $\checkmark$ & $\checkmark$ & $\checkmark$ &
\textbf{93.8} &
\textbf{69.6} &
95.6 &
\textbf{67.7} &
94.8 &
\textbf{85.1} &
\textbf{93.8} &
\textbf{85.8} &
\textbf{74.6} \\
\bottomrule
\end{tabular}
}
\end{table*}

\begin{table*}[h!]
\small
\centering
\caption{Quantitative comparison of different defect detection methods on the Wind Surface Defect dataset. Category-wise AP$_{50}$, overall mAP$_{50}$, mAP$_{50-95}$, and model parameters are reported. The top three results for each metric are highlighted in \TopOne{red}, \TopTwo{blue}, and \TopThree{green}, respectively.}
\label{tab:compare_wind_surface}
\resizebox{\textwidth}{!}{
  \begin{tabular}{lcccccccc}
  \toprule
  \multirow{2}{*}{Method} & \multicolumn{5}{c}{${\rm AP}_{50}$ (\%)} 
  & \multirow{2}{*}{${\rm mAP}_{50}$(\%)} 
  & \multirow{2}{*}{${\rm mAP}_{50-95}$(\%)} 
  & \multirow{2}{*}{Params} \\
  \cmidrule(lr){2-6}
   & surface-eye & corrosion & hide-craze & thunderstrike & dirt & & & \\
  \midrule
  RT-DETR-R34 &
  74.3 &
  61.3 &
  79.8 &
  76.2 &
  93.2 &
  76.9 &
  44.0 &
  31.1M \\

  RT-DETR-R50 &
  73.6 &
  61.8 &
  83.0 &
  \TopThree{81.3} &
  95.0 &
  78.9 &
  45.9 &
  41.6M \\

  MambaYOLO-B &
  \TopThree{79.0} &
  70.3 &
  \TopOne{89.2} &
  79.8 &
  \TopTwo{96.7} &
  \TopTwo{83.0} &
  \TopTwo{54.9} &
  21.8M \\

  DINO-4Scale-R50 &
  \TopOne{84.4} &
  64.0 &
  84.0 &
  79.4 &
  \TopThree{96.3} &
  \TopThree{81.6} &
  45.8 &
  47.2M \\

  YOLOv10-X &
  74.9 &
  62.8 &
  80.3 &
  66.5 &
  93.4 &
  75.6 &
  50.2 &
  31.6M \\

  YOLOv11-L &
  61.5 &
  \TopThree{72.1} &
  \TopThree{87.6} &
  81.0 &
  \TopOne{96.9} &
  79.8 &
  49.8 &
  25.3M \\

  YOLOv12-L &
  59.9 &
  \TopTwo{72.2} &
  84.5 &
  \TopTwo{83.1} &
  95.9 &
  79.1 &
  \TopThree{50.5} &
  26.4M \\

  BladeYOLO (Ours) &
  \TopTwo{81.0} &
  \TopOne{80.0} &
  \TopTwo{88.8} &
  \TopOne{87.0} &
  95.6 &
  \TopOne{86.5} &
  \TopOne{57.4} &
  43.8M \\

  \bottomrule
  \end{tabular}
}
\end{table*}

\subsection{Ablation Experiments}

To validate the effectiveness of each proposed module, we conduct ablation experiments on the WTBlade-Defect dataset under identical training and evaluation settings. As summarized in \autoref{tab:module_comparison}, YOLOv12-L is adopted as the baseline, and ablation variants are constructed by progressively introducing the DINOv3-initialized ViT backbone, the Style-Injector, the Detail-Enhanced Multi-scale Branch, and the Cross-Mamba module. The evaluation includes category-wise AP$_{50}$, together with the overall mAP$_{50}$ and mAP$_{50-95}$.

The baseline YOLOv12-L achieves 83.3\% mAP$_{50}$ and 71.9\% mAP$_{50-95}$. After replacing the original backbone with the DINOv3-initialized ViT, the mAP$_{50}$ increases from 83.3\% to 84.2\%, whereas the mAP$_{50-95}$ drops from 71.9\% to 66.6\%. This result suggests that although the pre-trained ViT backbone provides stronger semantic representations and improves detection under a relatively loose IoU criterion, its localization accuracy under stricter IoU thresholds remains insufficient, particularly for small defects with ambiguous boundaries.

After introducing the Style-Injector, the overall performance is further improved to 85.3\% mAP$_{50}$ and 68.2\% mAP$_{50-95}$. This indicates that explicitly modeling style-related appearance variations, such as illumination changes and surface appearance inconsistencies, helps improve robustness, especially for texture-sensitive categories such as \emph{crack} and \emph{dirt}. When the Detail-Enhanced Multi-scale Branch is further incorporated, the mAP$_{50}$ increases to 85.7\% and the mAP$_{50-95}$ recovers to 71.1\%. This demonstrates that explicitly introducing multi-scale detail-preserving pathways is beneficial for retaining fine-grained structural cues and improving localization performance for defects exhibiting large-scale variations.

Finally, by incorporating the Cross-Mamba module, the proposed method achieves the best performance, reaching 85.8\% mAP$_{50}$ and 74.6\% mAP$_{50-95}$. This result shows that progressively propagating high-level semantic guidance to shallow features can further strengthen defect awareness and localization accuracy across categories, especially for small-scale and easily confused defects.

\begin{figure*}[]
  \centering
  \includegraphics[width=1.0\textwidth]{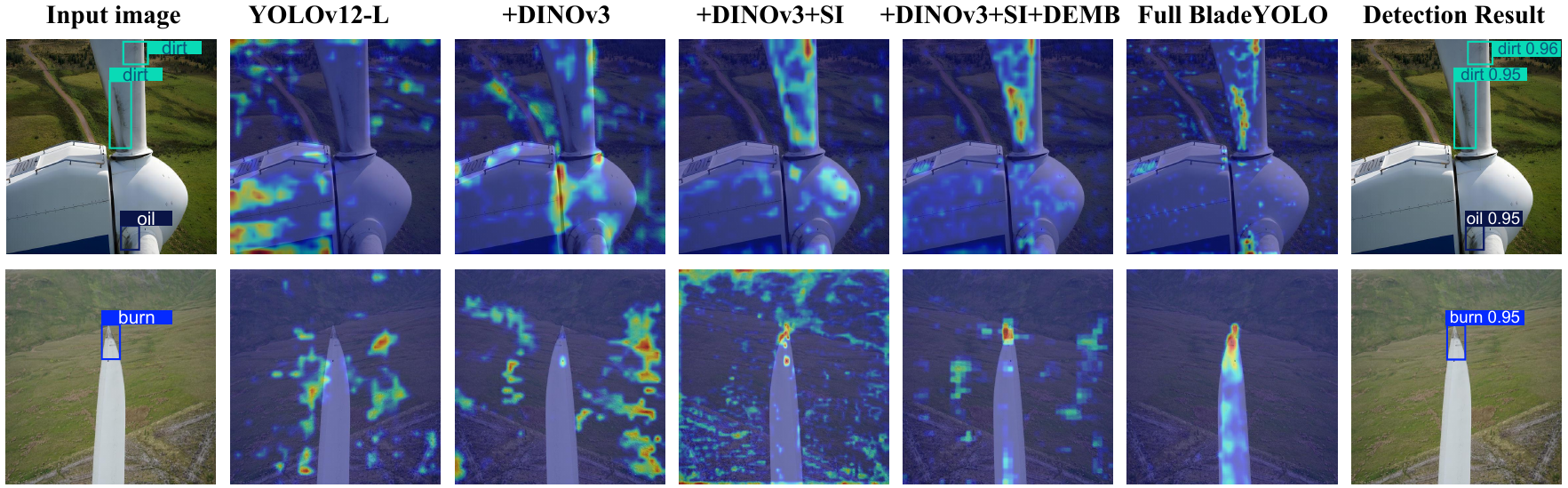}
  \caption{Heatmap-based ablation visualization of representative model variants on the WTBlade-Defect dataset. From left to right are the input image, YOLOv12-L baseline, +DINOv3, +DINOv3+SI, +DINOv3+SI+DEMB, the full BladeYOLO model (+DINOv3+SI+DEMB+Cross-Mamba), followed by the final detection results of BladeYOLO. SI denotes the Style-Injector, and DEMB denotes the Detail-Enhanced Multi-scale Branch. The visualizations illustrate that each module progressively enhances the model’s ability to focus on defect regions while suppressing background interference.}
  \label{fig:ablation_visualization}
  \vspace{-3mm}
\end{figure*}

\rev{
We further provide heatmap-based ablation visualizations in \autoref{fig:ablation_visualization} to illustrate the effect of each module. Compared with the baseline YOLOv12-L, the DINOv3-initialized ViT backbone produces stronger responses around defect regions, but the activations are still relatively scattered for small or ambiguous defects. The Style-Injector suppresses background interference and enhances the response strength in blade regions, thereby improving the model’s feature representations under varying appearance conditions. The Detail-Enhanced Multi-scale Branch further strengthens the responses around fine-grained structural cues and defect boundaries, making small defect regions more distinguishable. After introducing Cross-Mamba, the full BladeYOLO generates more concentrated and discriminative activations on defect regions while suppressing irrelevant background responses. The detection results in the rightmost column also show more accurate defect localization and classification, providing intuitive evidence for the complementarity of the proposed modules.
}

Overall, the ablation results demonstrate that each proposed module contributes positively to the final detection performance. More importantly, these modules exhibit strong complementarity. Compared with the baseline, the complete model achieves an improvement of 2.5\% in mAP$_{50}$ and 2.7\% in mAP$_{50-95}$, which validates the rationality and effectiveness of the proposed architecture for wind turbine blade defect detection under complex inspection conditions.

\subsection{Limited-Annotation Evaluation}

\rev{
To explicitly validate the effectiveness of BladeYOLO under limited-annotation conditions, we conduct annotation-budget experiments on the WTBlade-Defect dataset. Specifically, 10\%, 20\%, 50\%, and 70\% of the training images together with their annotations are sampled from the original training set using a fixed random seed, with the original category distribution approximately preserved, while 100\% of the training set is used as the full-annotation reference. The test set remains unchanged for fair evaluation. BladeYOLO is compared with representative YOLO-family baselines, including YOLOv8-L, YOLOv11-L, and YOLOv12-L. For each annotation ratio, all methods use the same sampled training subset, input resolution, optimizer, learning-rate schedule, batch size, and evaluation protocol.
}

\begin{table}[t]
\centering
\caption{\rev{Limited-annotation evaluation on the WTBlade-Defect dataset. Each cell reports mAP$_{50}$ / mAP$_{50-95}$ (\%).}}
\label{tab:limited_annotation}
\resizebox{\columnwidth}{!}{
\begin{tabular}{lccccc}
\hline
\diagbox[width=2.7cm,height=0.55cm]{Method}{Ratio} 
& 10\% & 20\% & 50\% & 70\% & 100\% \\
\hline
YOLOv8-L    & 42.1/21.5 & 60.2/37.2 & 63.4/43.2 & 75.9/63.2 & 85.1/74.4 \\
YOLOv11-L   & 42.8/20.2 & 59.3/33.2 & 65.6/36.8 & 72.1/60.0 & 84.6/73.9 \\
YOLOv12-L   & 45.8/21.1 & 58.5/32.5 & 64.1/36.6 & 71.7/58.7 & 83.3/71.9 \\
BladeYOLO (Ours) & 55.7/29.5 & 70.4/44.5 & 72.1/49.5 & 79.3/64.0 & 85.8/74.6 \\
\hline
\end{tabular}
}
\vspace{-3mm}
\end{table}

\rev{
As shown in \autoref{tab:limited_annotation}, BladeYOLO consistently achieves the best performance across all annotation ratios. The advantage is particularly evident when the annotation ratio is low. Under the 10\% annotation setting, BladeYOLO achieves 55.7\% mAP$_{50}$ and 29.5\% mAP$_{50-95}$, outperforming the direct YOLOv12-L baseline by 9.9 and 8.4 percentage points, respectively. Under the 20\% annotation setting, BladeYOLO improves mAP$_{50}$ and mAP$_{50-95}$ by 11.9 and 12.0 percentage points over YOLOv12-L, respectively. These results indicate that the DINOv3-initialized ViT backbone provides transferable representation priors, and the proposed adaptation modules further help exploit these priors for blade defect detection under scarce task-specific annotations.
}

\rev{
When more annotations are available, BladeYOLO still maintains competitive advantages. For example, under the 50\% annotation setting, BladeYOLO improves mAP$_{50}$ and mAP$_{50-95}$ over YOLOv12-L by 8.0 and 12.9 percentage points, respectively. Under the full-annotation setting, the performance gap becomes smaller, but BladeYOLO still achieves the highest mAP$_{50}$ and mAP$_{50-95}$. These observations suggest that the proposed framework shows more pronounced advantages under limited-annotation conditions, while also remaining effective when sufficient annotations are available.
}

\subsection{Generalization Evaluation}
To further evaluate the robustness and generalization ability of the proposed method, additional experiments are conducted on the publicly available Wind Surface Defect dataset~\cite{Liu2025Wind}. On this dataset, we retrain \emph{BladeYOLO} and compare it with several representative methods, including RT-DETR-R50, YOLOv11-L, YOLOv12-L, and MambaYOLO-B, under the same experimental settings used on the WTBlade-Defect dataset. The quantitative results are summarized in \autoref{tab:compare_wind_surface}. To provide a more intuitive comparison of category-level performance, \autoref{fig:radar_sci}(b) further presents the radar chart of category-wise AP$_{50}$ together with the overall mAP$_{50}$ on the Wind Surface Defect dataset.

As shown in \autoref{tab:compare_wind_surface} and \autoref{fig:radar_sci}(b), the proposed method achieves the best overall performance, reaching 86.5\% mAP$_{50}$ and 57.4\% mAP$_{50-95}$. \rev{The radar chart further shows that \emph{BladeYOLO} achieves balanced and competitive performance across the five defect categories, suggesting its promising generalization performance on the public Wind Surface Defect dataset.} In particular, the proposed method exhibits clear advantages on visually subtle defect categories such as \emph{hide-craze} and \emph{surface-eye}, which are especially challenging due to slight texture differences and complex backgrounds. These results further verify the robustness of the proposed method under diverse surface appearances and background interference.
\rev{Overall, the proposed method achieves competitive performance on wind turbine blade defect detection and shows promising generalization performance on a related public surface defect dataset, highlighting its potential for broader industrial inspection applications.}

\subsection{Exploratory Qualitative Analysis with SAM 3}

\rev{
To explore the potential of prompt-based foundation models for future wind turbine blade defect segmentation, we conduct an exploratory qualitative analysis using SAM 3~\cite{carion2025sam}. This analysis is not intended to serve as a primary baseline comparison for validating the proposed detection architecture. Since pixel-level defect mask annotations are difficult to obtain and are unavailable in our datasets, SAM 3 is used without task-specific retraining or mask-level supervision. Therefore, we do not conduct quantitative comparisons with SAM 3. Instead, we use it as a qualitative probe to observe how a prompt-based foundation model responds to blade-surface defects under complex inspection conditions, thereby providing preliminary insights for future mask-level blade defect analysis.
}

\begin{figure}[tbp]
  \centering
  \includegraphics[width=1.0\linewidth]{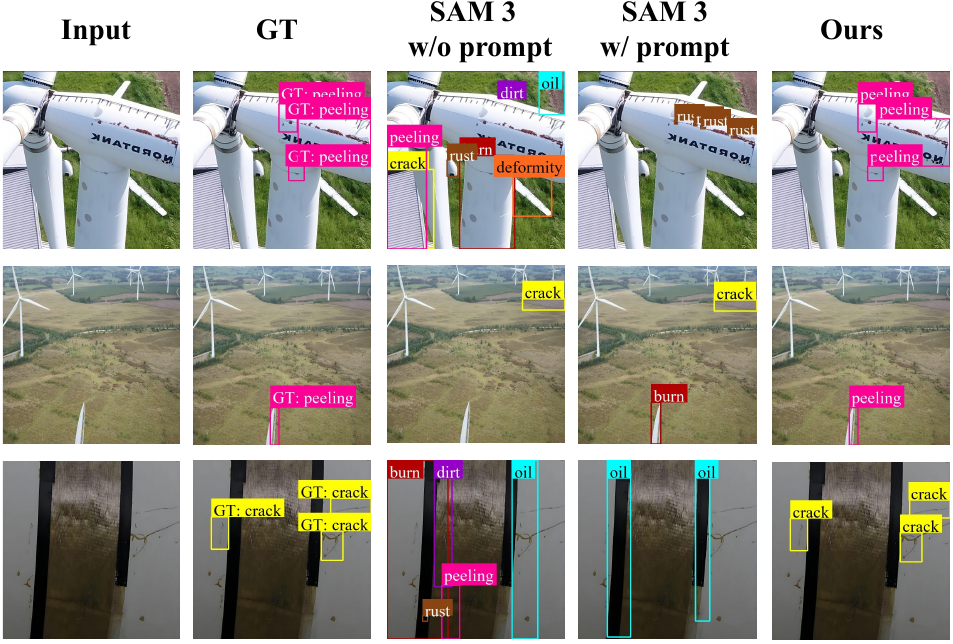}
  \caption{\rev{
Exploratory qualitative analysis of SAM 3 and \emph{BladeYOLO}. From left to right are the input image, ground truth, SAM 3 without prompts, SAM 3 with expert prompts, and \emph{BladeYOLO}. Expert prompts help SAM 3 focus more on defect-related regions and reduce redundant detections, while \emph{BladeYOLO} produces tighter predictions that are more consistent with the ground-truth bounding boxes. This analysis is used as a qualitative probe to examine the potential of prompt-based foundation models for future blade defect segmentation, rather than as a primary supervised baseline comparison.
}}
  \label{fig:fig_sam}
\end{figure}

Given the zero-shot nature of SAM 3, we adopt an expert prompting strategy by providing descriptive textual cues, such as ``thin dark linear cracks'' and ``grayish dust, environmental dirt, or sand accumulation,'' to guide defect localization. As illustrated in \autoref{fig:fig_sam}, semantic prompting has a clear influence on the behavior of SAM 3. Without prompts, SAM 3 tends to generate redundant bounding boxes and incorrect category assignments, mainly because it lacks task-specific priors for blade-surface defects. After introducing expert prompts as semantic anchors, SAM 3 is better constrained to regions that are more consistent with the visual and physical characteristics of defects, which helps reduce background-induced false responses and implausible predictions.

Nevertheless, even with expert guidance, SAM 3 still exhibits localization deviations and category confusion on complex blade surfaces. These observations indicate that directly applying prompt-based foundation models to blade defect analysis remains challenging without task-specific adaptation and pixel-level supervision.
\rev{
Overall, this exploratory analysis suggests that SAM 3 may provide useful qualitative cues for defect localization and has potential for future blade defect segmentation research. 
}

\section{Conclusion}
\label{sec:conclusion}
In this paper, we proposed \emph{BladeYOLO}, a hybrid detection framework for wind turbine blade defect detection. By integrating a DINOv3-initialized ViT into YOLOv12-L and further introducing the Style-Injector, the Detail-Enhanced Multi-scale Branch, and the Cross-Mamba module, the proposed method effectively addresses several key challenges in real-world blade inspection, including limited training data, environment-induced appearance variations, and the weak visual saliency of small-scale defects. Hence, \emph{BladeYOLO} achieves enhanced robustness and accuracy under complex inspection conditions.

Extensive experiments on the WTBlade-Defect dataset demonstrate the effectiveness of the proposed framework. \emph{BladeYOLO} achieves 85.8\% mAP$_{50}$ and 74.6\% mAP$_{50-95}$, outperforming several representative detection methods. These results demonstrate that the proposed method can effectively improve defect detection performance in industrial scenarios, where image quality is often affected by varying illumination, foggy or hazy conditions, and background interference.

To further evaluate its generalization ability, we also conducted experiments on the public Wind Surface Defect dataset. The proposed method achieves 86.5\% mAP$_{50}$ and 57.4\% mAP$_{50-95}$, outperforming the competing methods. Despite the differences in data distribution and defect characteristics between the two datasets, \emph{BladeYOLO} consistently maintains competitive detection performance, showing its robustness and adaptability across related surface defect detection scenarios.

Overall, \emph{BladeYOLO} provides an effective and practical solution for automated wind turbine blade inspection, achieving a favorable balance among detection accuracy, robustness to environmental interference, and model complexity. 
\rev{Therefore, the proposed method provides support for more efficient automated detection and maintenance of wind turbine blades.}
In future work, we will focus on improving its efficiency for real-time and resource-constrained deployment. In particular, we plan to explore lightweight architectural design, structured pruning, and knowledge distillation to further reduce the computational cost. We will also investigate deployment-oriented acceleration strategies, such as ONNX-to-TensorRT optimization, to facilitate real-time inference on edge devices and unmanned aerial vehicles, supporting automated inspection.

\section{Acknowledge}
This work was supported by the Zhejiang Provincial Major Science and Technology Project (No. 2026LDC01046(GZ)), Zhejiang Provincial Natural Science Foundation of China under Grant No. LQ23F020023, and the Science Foundation of Zhejiang Sci-Tech University(ZSTU) (No.22242257-Y)
.

\bibliographystyle{IEEEtran}
\bibliography{elsarticle} 

\end{document}